\documentclass[10pt,twocolumn,letterpaper]{article}
\pdfoutput=1
\usepackage{iccv}
\usepackage{times}
\usepackage{epsfig}
\usepackage{graphicx}
\usepackage{amsmath}
\usepackage{amssymb}
\usepackage{multirow}
\usepackage{comment}
\usepackage[export]{adjustbox}
\usepackage{stfloats}
\usepackage[numbers]{natbib}
\usepackage{setspace}
\usepackage{hyperref}
\hypersetup{colorlinks = true,  linkcolor =red,  anchorcolor = red,  urlcolor = magenta}
\usepackage[accsupp]{axessibility}

 \iccvfinalcopy 

\def\iccvPaperID{4779} 

\ificcvfinal\fi

\begin{document}
\setlength{\lineskiplimit}{0pt}
\setlength{\lineskip}{0pt}
\setlength{\abovedisplayskip}{3pt}   
\setlength{\belowdisplayskip}{3pt}
\setlength{\abovedisplayshortskip}{3pt}
\setlength{\belowdisplayshortskip}{3pt}

\title{GridMM: Grid Memory Map for Vision-and-Language Navigation}

\author{Zihan Wang$^{1,2}$, Xiangyang Li$^{1,2}$, Jiahao Yang$^{1,2}$, Yeqi Liu$^{1,2}$, Shuqiang Jiang$^{1,2}$\\
$^{1}$Key Lab of Intelligent Information Processing Laboratory of the Chinese Academy of Sciences (CAS),\\
Institute of Computing Technology, Chinese Academy of Sciences, Beijing, 100190, China\\
$^{2}$University of Chinese Academy of Sciences, Beijing, 100049, China\\
\tt\small zihan.wang@vipl.ict.ac.cn, lixiangyang@ict.ac.cn,\\
\tt\small \{jiahao.yang, yeqi.liu\}@vipl.ict.ac.cn, sqjiang@ict.ac.cn
}

\maketitle

\ificcvfinal\thispagestyle{empty}\fi


\begin{abstract}

Vision-and-language navigation (VLN) enables the agent to navigate to a remote location following the natural language instruction in 3D environments. To represent the previously visited environment, most approaches for VLN implement memory using recurrent states, topological maps, or top-down semantic maps. In contrast to these approaches, we build the top-down egocentric and dynamically growing Grid Memory Map (\textit{i.e.}, GridMM) to structure the visited environment. From a global perspective, historical observations are projected into a unified grid map in a top-down view, which can better represent the spatial relations of the environment. From a local perspective, we further propose an instruction relevance aggregation method to capture fine-grained visual clues in each grid region. Extensive experiments are conducted on both the REVERIE, R2R, SOON datasets in the discrete environments, and the R2R-CE dataset in the continuous environments, showing the superiority of our proposed method. The source code is available at
\href{https://github.com/MrZihan/GridMM}{https://github.com/MrZihan/GridMM}. 
\end{abstract}

\section{Introduction}\label{introduction}

Vision-and-language navigation (VLN) tasks\ \cite{VLN-2018vision,2020-RXR,2020reverie} 
 require an agent to understand
natural language instructions and act according to the instructions. Two
distinct VLN scenarios have been proposed, being navigation in discrete
environments (\textit{e.g.}, R2R\ \cite{VLN-2018vision}, REVERIE\ \cite{2020reverie}, SOON\ \cite{zhu2021soon}) and in continuous environments
(\textit{e.g.}, R2R-CE\ \cite{Krantz2020r2r-ce}, RxR-CE\ \cite{2020-RXR}). The discrete environment in VLN
is abstracted as the topology structure of interconnected navigable nodes.
With the connectivity graph, the agent can move to an adjacent node
on the graph by selecting a direction from navigable directions.
Different from the discrete environments, VLN in continuous environments require the agent to move through low-level controls (\textit{i.e.}, turn left 15 degrees, turn right 15 degrees, or move forward 0.25
meters), which is closer to real-world robot navigation and more challenging.

\begin{figure}
\noindent\begin{minipage}[t]{1\columnwidth}%
\begin{center}
\includegraphics[width=1\columnwidth]{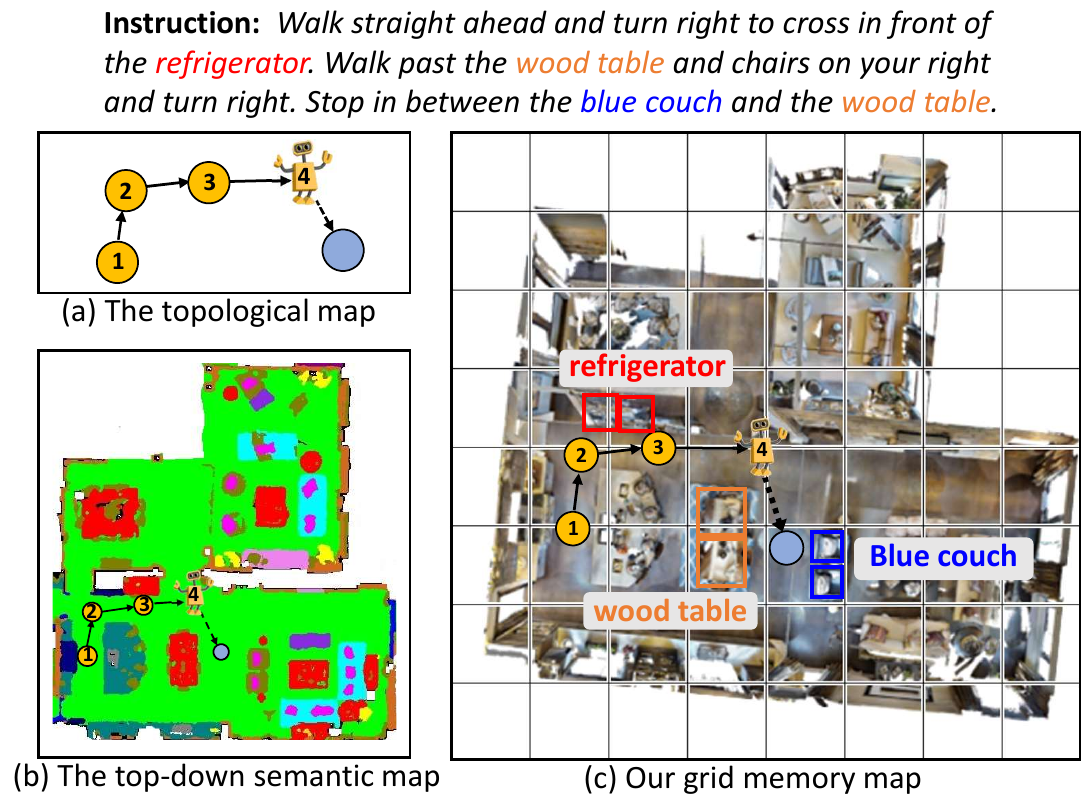}
\par\end{center}%
\end{minipage}
\caption{Illustration of different methods to represent the environment with maps for VLN. 
}
\label{fig:introduction}
\end{figure}

Whether in discrete environments or continuous environments,
historical information during navigation plays an important role
in environment understanding and instruction grounding. In previous
works\ \cite{VLN-2018vision,2018-speaker,tan2019learning,2019reinforced,hong2021vln-bert}, recurrent states are most commonly used as historical information
for VLN, which encode historical observations and actions within a
fixed-size state vector. However, such condensed states might be insufficient for capturing essential information in trajectory history. Therefore, Episodic transformer~\cite{pashevich2021episodic} and HAMT~\cite{chen2021history} propose to 
 directly encode the trajectory
history and actions as a sequence of previous observations instead of using recurrent states. Furthermore, in order to structure the
visited environment and make global planning, a few recent approaches~\cite{chen2021topo,chen2022think-GL, VLN_LAD_2023} structure
the topological map, as shown in Fig.~\ref{fig:introduction}(a). However, these methods are difficult to represent the spatial relations among objects and scenes in historical observations, 
thus a lot of detailed information is lost. 
As shown in Fig.~\ref{fig:introduction}(b), more recent works~\cite{SASRA,georgakis2022cm2,chen2022weakly,huang23vlmaps} model the navigation environment using the top-down semantic map, which represents spatial relations more precisely. 
But the semantic concepts are extremely limited due to the pre-defined semantic labels. So the objects or scenes, which are not included in prior semantic labels, cannot be represented, such as the “refrigerator” in Fig.~\ref{fig:introduction}(b). Moreover, 
as illustrated in Fig.~\ref{fig:introduction}(b), the objects with diverse attributes such as “wood table” and “blue couch” cannot be fully expressed by the semantic map which misses object attributes. 

In contrast to the above works~\cite{chen2021topo,chen2022think-GL,georgakis2022cm2,huang23vlmaps}, we propose the Grid Memory
Map (\textit{i.e.}, GridMM), a visual representation structure for modeling global historical observations during navigation. 
Different from BEVBert~\cite{Dong2022bevbert}, who  applies local hybrid metric maps for short-term reasoning, our GridMM leverages both temporal and spatial information to depict the globally visited environment. Specifically, the grid map divides the visited environment into many equally large grid regions, and each grid region contains many fine-grained visual features. We dynamically construct a grid memory bank to update the grid map during navigation. At each step of navigation, the visual features from the pre-trained CLIP~\cite{radford2021learning} model are saved into the memory bank, and all of them are categorized into the grid map regions based on their coordinates calculated via the depth information. To obtain the representation of each region, we design an instruction relevance aggregation method to capture the visual features most relevant to instructions and aggregate them into one holistic feature. With the help of $N$$\times$$N$ aggregated map features, the agent is able to accurately conduct the next action planning. A wealth of experiments illustrate the effectiveness of our GridMM compared with the previous methods.

In summary, we make the following contributions:
\begin{itemize}
\vspace{-0.2cm}
\item We propose the Grid Memory Map for VLN to structure the global space-time relations of the visited environment and adopt instruction relevance aggregation to capture visual clues relevant to instructions. 

\vspace{-0.2cm}
\item We 
comprehensively compare different 
maps representing the visited environment in VLN and analyze the 
characteristics of our proposed 
GridMM, which depicts more fine-grained information and gives some insights into future works in VLN.

\vspace{-0.2cm}
\item Extensive experiments are conducted to verify the effectiveness of our method in both discrete environments and continuous environments, which show that our
method outperforms existing methods on many benchmark datasets.
\end{itemize}

\section{Related work}

\noindent \textbf{Vision-and-Language Navigation (VLN).} VLN\ \cite{VLN-2018vision,2019reinforced,hong2020l-e-graph,2021structured-scene,qiao2022HOP,chen2022think-GL, Chen_2022_HM3D_AutoVLN} has received significant attention in recent years with the continual
improvement. 
The VLN tasks include 
step-by-step instructions such as R2R\ \cite{VLN-2018vision} and RxR \ \cite{2020-RXR}, navigation with dialog
such as CVDN\ \cite{thomason2020cvdn}, and navigation for remote object grounding
such as REVERIE\ \cite{2020reverie} and SOON\ \cite{zhu2021soon}. All tasks require the agent's ability to use time-dependent
visual observations for decision-making. Restricted by the heavy computation
of exploring the large action space in continuous environments, early works mainly focused on discrete environments. Among them,
a recurrent unit is usually utilized to encode historical
observations and actions within a fixed-size state vector \ \cite{VLN-2018vision,2018-speaker,tan2019learning,2019reinforced,hong2021vln-bert}.
Instead of relying on the recurrent states, HAMT\ \cite{chen2021history} explicitly encodes
the panoramic observation history to capture long-range dependency,
and DUET\ \cite{chen2022think-GL} proposes to encode the topological map for efficient global planning. Inspired by the success of vision-and-language
pre-training\ \cite{2019-lxmert,radford2021learning}, HOP\ \cite{qiao2022HOP,qiao2023hop+} utilizes well-designed proxy tasks for pre-training to enhance
the interaction between vision and language modalities. ADAPT~\cite{lin2022adapt} employs action prompts to improve the cross-modal alignment ability. Based on data augmentation methods, some
approaches enlarge training data of visual modality\ \cite{li2022envedit} and linguistic
modality\ \cite{2018-speaker,liang2022visual,dou2022foam,Scaling_rxr} depending on existing VLN datasets. Moreover, AirBERT\ \cite{2021airbert} and
HM3D-AutoVLN\ \cite{Chen_2022_HM3D_AutoVLN} improve the performance by creating large-scale
training dataset. KERM~\cite{Li2023KERM} utilizes a large knowledge base to depict navigation views for better generalization ability. In this work, we propose a dynamically growing grid memory map for structuring the visited environment and making long-term planning, which facilitates environment understanding and instruction grounding.

\noindent \textbf{VLN in Continuous Environments (VLN-CE).} VLN-CE~\cite{Krantz2020r2r-ce}
converts the topologically-defined VLN
tasks such as R2R\ \cite{VLN-2018vision} into the continuous environment tasks, which
is closer to real-world navigation. Different from the discrete environments,
the agent in VLN-CE must navigate to the destination by selecting low-level
action, similar to some visual navigation tasks~\cite{zsx_iccv,zsx_eccv,isia_mm,Zhang_cvpr,tang2022monocular,tang2021auto,zhu2019sim}. Some approaches~\cite{georgakis2022cm2,chen2022weakly} apply top-down semantic maps for environment understanding and use language-aligned waypoints supervision~\cite{SASRA} for action prediction. Recently, Bridging~\cite{Hong2022bridging} and Sim-2-Sim~\cite{krantz2022sim2sim} for transferring pre-trained VLN
agents to continuous environments have achieved considerable results.
Compared with training agents from scratch in VLN-CE,
this strategy can reduce the computational cost of pre-training
and accelerate model convergence.
In this work, we pre-train our model based on the proposed GridMM in discrete environments
and then transfer the 
model to continuous environments. Experiments in both discrete environments and continuous environments illustrate the effectiveness of our method.

\begin{figure*}[htbp]
\makebox[\textwidth][c]
{\includegraphics[width=0.74\paperwidth]{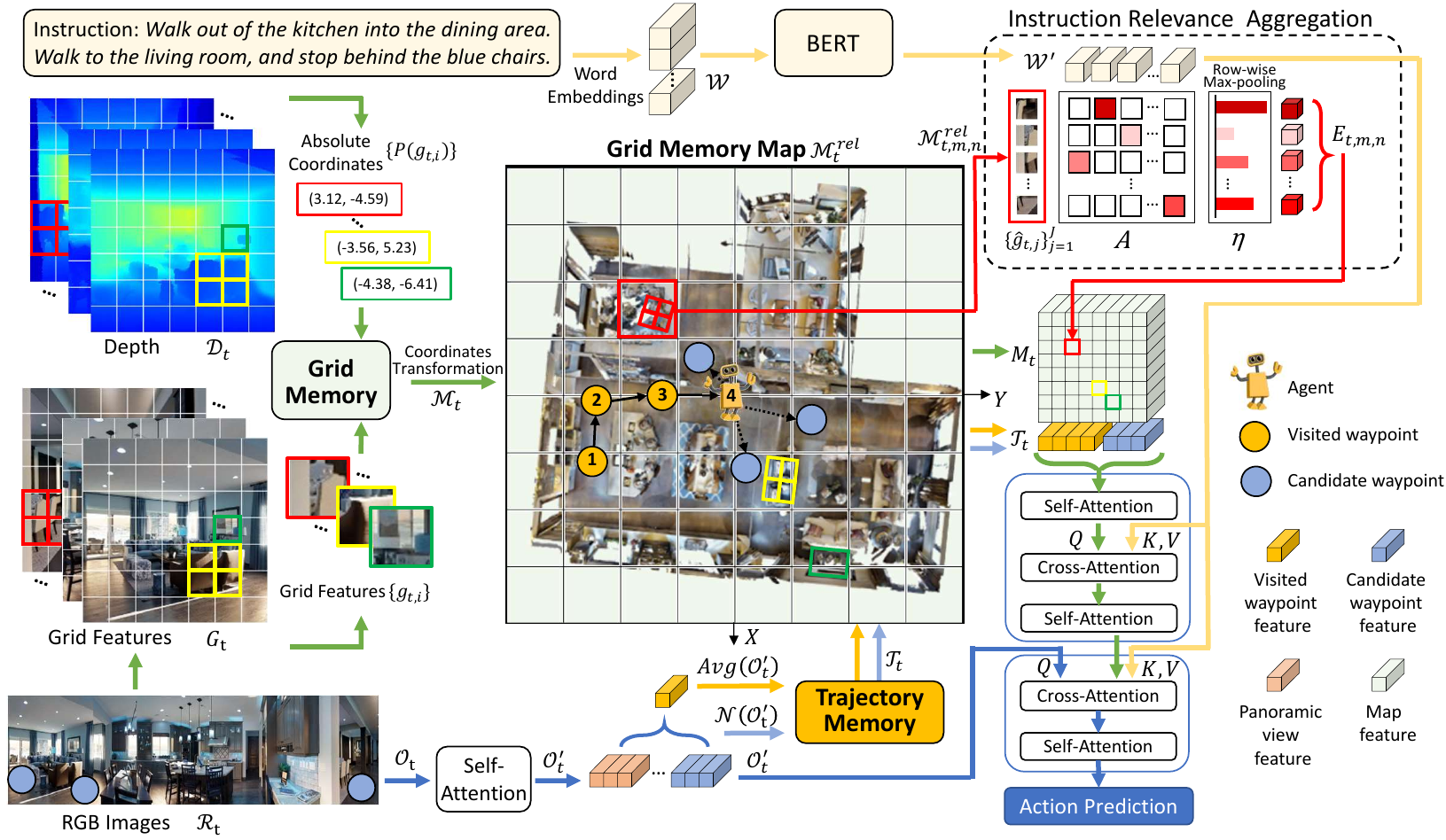}}\caption{The overall pipeline. At step $t$, fine-grained grid features $G_t$ are extracted from panoramic observations $\mathcal{R}_{t}$ and stored into grid memory $\mathcal{M}_{t}$ with their absolute coordinates (calculated via depth images $\mathcal{D}_{t}$ and coordinate of the agent). Waypoints of trajectory are represented via panoramic features $\mathcal{O}_{t}^{'}$, and then stored into trajectory memory $\mathcal{T}_{t}$. An egocentric grid memory map can be constructed by projecting all features of $\mathcal{M}_{t}$ into a unified square map with $N$$\times$$N$ cells. Map features can be obtained by aggregating all features in each cell with an instruction relevance method. Panoramic view features, map features, and trajectory features are fed into a two-stage cross-modal encoder for action reasoning. For simplicity, layer normalization, feed-forward network, and residual structure are omitted from this figure. Best viewed in color.}
\label{fig:framework}
\end{figure*}

\noindent \textbf{Maps for Navigation.} The works on visual navigation~\cite{gupta2017cognitive,chaplot2020object,wani2020multion} and other 3D indoor scene understanding tasks~\cite{henriques2018mapnet,beeching2020egomap,mapnet_aaai,Datta_2022_CVPR}
has a long tradition of constructing maps. Some works represent the
map as topological structures for back-tracking to other locations~\cite{chen2021topo}
or supporting global action planning~\cite{chen2022think-GL}. In addition, some approaches\ \cite{georgakis2022cm2,huang23vlmaps}
construct a top-down semantic map to more precisely represent spatial relations of the environment. 
Recently, BEVBert\ \cite{Dong2022bevbert}  introduced topo-metric maps from robotics into VLN, which uses topological maps for long-term planning and applies hybrid metric maps for short-term reasoning. 
Its metric map divides the local environment around the agent into $21$$\times$$21$ cells, and each cell represents a square region with a side length of $0.5$$m$. Moreover, the short-term visual observations within two steps are mapped into these cells. However, our GridMM is completely different in terms of: \textbf{(1)} 
BEVBert enriches the representations of the local observation with grid features. 
Our GridMM aims to perceive more space-time relationships with the dynamically growing grid map, which leverages both temporal and spatial information to depict the globally visited environment.  
 \textbf{(2)} The grid-based metric map in BEVBert is only used for local action prediction. Our GridMM expands with the expansion of the visited environment, providing spatial enhanced representations for both local and global action prediction.
\textbf{(3)} The representations of the metric map for BEVBert are only the visual features. The representations of each cell in our GridMM are self-adapted  to the instructions, which contain both visual and linguistic  information.

\section{Method}

\subsection{Navigation Setups}\label{sec:navigation_setups}

For VLN in discrete environments, the navigation connectivity graph\ $\mathcal{G}=\{\mathcal{V},\mathcal{E}\}$ is provided in the Matterport3D simulator\ \cite{matterport3d}, where\ $\mathcal{V}$ denotes navigable
nodes and\ $\mathcal{E}$ denotes edges. An agent is equipped with
RGB and depth cameras, and a GPS sensor. Initialized at a starting
node and given natural language instructions, the agent needs to explore
the navigation connectivity graph $\mathcal{G}$ and reach the target
node. \ $\mathcal{W}=\{w_{l}\}_{l=1}^{L}$ denote the word embeddings of the instruction with $L$ words. At each time step $t$, the agent
observes panoramic RGB images $\mathcal{R}_{t}=\{r_{t,k}\}_{k=1}^{K}$ and the depth images $\mathcal{D}_{t}=\{d_{t,k}\}_{k=1}^{K}$ of its current node $\mathcal{V}_{t}$, which contains $K$ single
view images. The agent is aware of a few navigable
views $\mathcal{N}(\mathcal{R}_{t})\in\mathcal{R}_{t}$ corresponding to its neighboring nodes and their coordinates. 

VLN in continuous environments is established over Habitat\ \cite{2019habitat}, where the agent's position $\mathcal{P}_{t}$ can be any point in the open space. In each navigation step, we use a pre-trained waypoint predictor\ \cite{Hong2022bridging} to generate navigable waypoints in continuous environments, which assimilates the task with the VLN in discrete environments.

\subsection{Grid Memory Mapping}\label{sec:grid_memory_mapping}

As illustrated in Fig.\ \ref{fig:framework}, we present our grid memory mapping pipeline. At each navigation step $t$, we first store the fine-grained visual features and their corresponding coordinates in the grid memory. For the panoramic RGB images $\mathcal{R}_{t}=\{r_{t,k}\}_{k=1}^{K}$, we use a pre-trained CLIP-ViT-B/32~\cite{radford2021learning} model to extract grid features $G_{t}=\{g_{t,k}\in\mathcal{\mathbb{R}}^{H\times W\times D}\}_{k=1}^{K}$, and the grid feature of row $h$ column $w$ is denoted as $g_{t,k,h,w}\in {\mathbb{R}}^{D}$. The corresponding depth images $\mathcal{D}_{t}$ are downsized to the same scale as $\mathcal{D}_{t}^{'}=\{d_{t,k}^{'}\in\mathcal{\mathbb{R}}^{H\times W}\}_{k=1}^{K}$, and the depth value of row $h$ column $w$ is denoted as $d_{t,k,h,w}^{'}$. For convenience, we denote all the subscripts $(k,h,w)$ as $i$, where $i$ ranges from 1 to $I$, and $I=K$$\cdot$$H$$\cdot$$W$. So $g_{t,k,h,w}$ is denoted as $\hat{g}_{t,i}$, and $d_{t,k,h,w}$ is denoted as $\hat{d}_{t,i}$. Similar to \cite{vln-pano2real,huang23vlmaps}, we can calculate the absolute coordinates $P(\hat{g}_{t,i})$ of $\hat{g}_{t,i}$:
\begin{align}
&P(\hat{g}_{t,i})=(x_{t,i}\ ,\ y_{t,i})\\&=(\mathcal{X}_{t}+d_{t,i}^{line}\cdot cos\theta_{t,i}\ ,\ \mathcal{Y}_{t}+d_{t,i}^{line}\cdot sin\theta_{t,i})\notag
\end{align}
where $(\mathcal{X}_{t},\mathcal{Y}_{t})$ denotes the agent's current coordinate, $\theta_{t,i}$ denotes the heading angle between $\hat{g}_{t,i}$ and the current orientation of agent, $d_{t,i}^{line}$ denotes the euclidean distance between $\hat{g}_{t,i}$ and the agent, which can be calculated via $\hat{d}_{t,i}$ and $\theta_{t,i}$. We store all these grid features and their absolute coordinates in the grid memory:
\begin{align}
\mathcal{M}_{t}= \mathcal{M}_{t-1} \cup \{[\hat{g}_{t,i}, P(\hat{g}_{t,i})]\}_{i=1}^{I}
\end{align}

\begin{figure}
\noindent\begin{minipage}[t]{1\columnwidth}%
\begin{center}
\includegraphics[width=1\columnwidth]{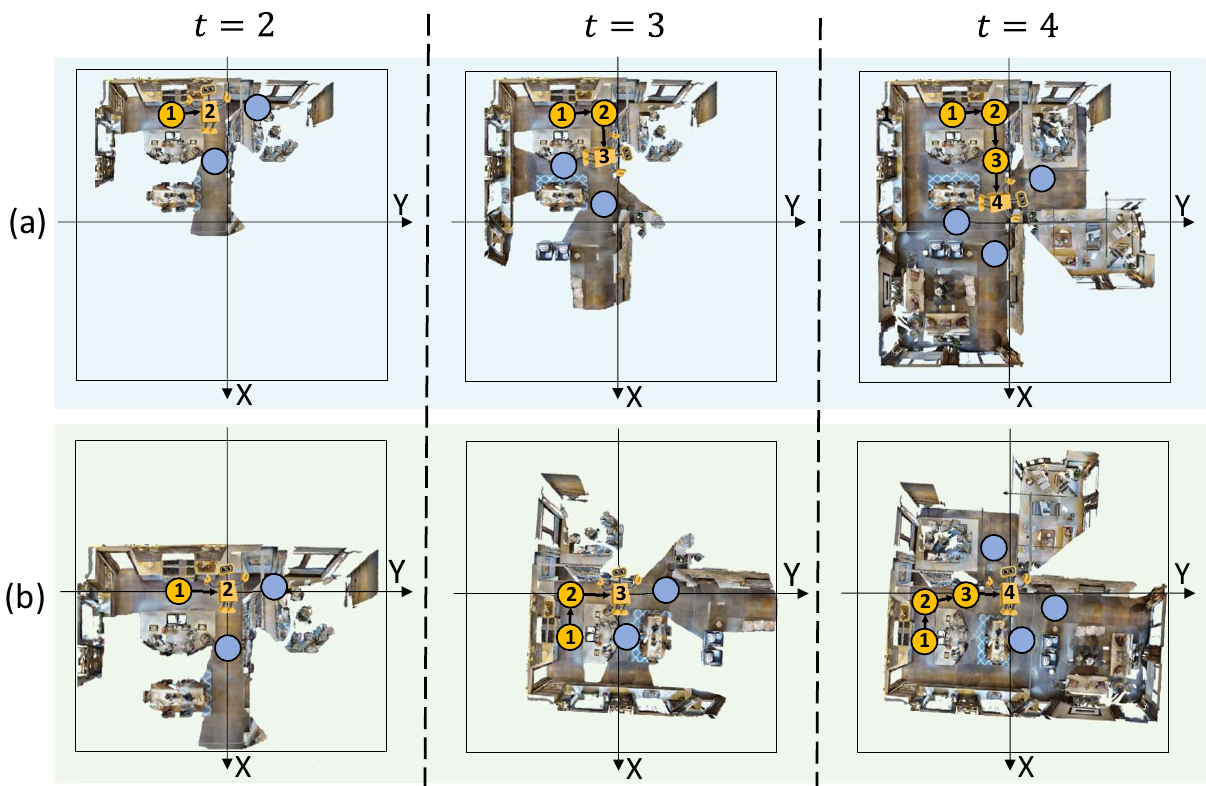}
\par\end{center}%
\end{minipage}
\caption{Maps in (a) use the absolute coordinate with a constant side length and coordinate origin. Maps in (b) use dynamically relative coordinates that the side length increases with the expansion of the visited environment, taking the position of the current agent as the coordinate origin and the direction of the current agent as the positive direction of the y-axis. Please zoom in for best view.}
\label{fig:map_grow}
\end{figure}

Then we propose a dynamic coordinate transformation method for constructing the grid memory map using visual features in grid memory $\mathcal{M}_{t}$. Intuitively, we can construct the maps as shown in Fig.\ \ref{fig:map_grow}(a). The visited environment is represented by projecting all historical observations $\hat{g}_{t,i}$ into unified maps based on their absolute coordinates $P(\hat{g}_{t,i})$. However, such maps have two drawbacks. First, it is not 
efficient enough to align the candidate observations and the instruction with the absolute coordinate. Second, it is difficult to determine the scale and extent of the map without prior information about the environment~\cite{zhao2022target}. 

To address these deficiencies, we propose a new mapping method to construct the top-down egocentric and dynamically growing map, as illustrated in Fig.\ \ref{fig:map_grow}(b). At each step, we build a grid map in an egocentric view by projecting all features of the grid memory $\mathcal{M}_{t}$ into a new planar cartesian coordinate system with the agent's position as the coordinate origin and the agent's current direction as the positive direction of the y-axis.
In this new coordinate system, for each grid feature $\hat{g}_{s,i}$ in $\mathcal{M}_{t}$ (where $s$ ranges from 1 to $t$),  we can calculate the new relative coordinates $P_t^{rel}(\hat{g}_{s,i})$ in time step $t$:
\begin{align}
&P_t^{rel}(\hat{g}_{s,i})=(x_{s,i}^{rel}\ ,\ y_{s,i}^{rel})\notag\\&=(\ ({x}_{s,i}-\mathcal{X}_{t})\cdot cos\Theta_{t}+({y}_{s,i}-\mathcal{Y}_{t})\cdot sin\Theta_{t}\ ,\ \notag\\&\ \ \ \ \ \ \ \ ({y}_{s,i}-\mathcal{Y}_{t})\cdot cos\Theta_{t}-({x}_{s,i}-\mathcal{X}_{t})\cdot sin\Theta_{t}\ )\label{formula:relative_coordinates}
\end{align}
where $\Theta_{t}$ represents the heading angle between the new coordinate system and the old coordinate system. 

Further, we construct the grid memory map (\textit{i.e.}, GridMM) via the grid features and their new coordinates. 
At step $t$, the grid memory map takes $L_{t}$ as the side length: 
\begin{align}
\label{formula:side_length}
\small
L_{t}=2\cdot max(\ max(\{\{|x_{s,i}^{rel}|\}_{i=1}^{I}\}_{s=1}^{t})\ ,\\ \notag max(\{\{|y_{s,i}^{rel}|\}_{i=1}^{I}\}_{s=1}^{t})\ )
\end{align}
So that the size of the GridMM increases with the expansion of the visited environment. The agent is always in the center of this map and the map is aligned with current panoramic observations in an egocentric view. Then the map is divided into $N$$\times$$N$ cells and all features of $\mathcal{M}_{t}$ are projected into these cells according to their new relative coordinates.  Finally, we
construct the grid memory map $\mathcal{M}_{t}^{rel}$ with $N$$\times$$N$ cells, and each cell contains multiple fine-grained visual features. After aggregating all visual features in each cell into one embedding vector, the map features $M_{t}\in\mathcal{\mathbb{R}}^{N\times N \times D}$ is obtained. The detailed aggregation method is described in Sec.\ \ref{sec:instruction relevance_grid_memory_encoding}.

\subsection{Model Architecture}

\subsubsection{Instruction and Observation Encoding}\label{rec:observation_encoding}

For instruction encoding, each word embedding in $\mathcal{W}$ is added with a position embedding
and a token type embedding. All tokens are then fed into a multi-layer
transformer to obtain word representations, denoted as $\mathcal{W}^{'}=\{w_{l}^{'}\}_{l=1}^{L}$.

For view images $\mathcal{R}_{t}$ of the panoramic observation, we use the ViT-B/16\ \cite{2020image-vit} pre-trained on ImageNet to extract visual features $\mathcal{R}_{t}^{'}$.
Then we represent their relative angles as $a_{t}=(sin\theta_{t}^a,cos\theta_{t}^a,sin\varphi_{t}^a,cos\varphi_{t}^a)$, where $\theta_{t}^a$ and $\varphi_{t}^a$ are the relative heading
and elevation angles to the agent's orientation. The candidate waypoints are represented as $\mathcal{N}(\mathcal{R}_{t}^{'})$, and the line distance between waypoints and the current agent is denoted as
$b_{t}$. Similarly, we represent the relative angles between the agent
and the start waypoint as $c_{t}=(sin\theta_{t}^c,cos\theta_{t}^c,sin\varphi_{t}^c,cos\varphi_{t}^c)$.
Then we concatenate
the line distance $dist_{line}(\mathcal{V}_{0},\mathcal{V}_{t})$,
navigation trajectory length $dist_{traj}(\mathcal{V}_{0},\mathcal{V}_{t})$,
and action step $dist_{step}(\mathcal{V}_{0},\mathcal{V}_{t})$
between agent and the start waypoint to obtain $e_{t}=(dist_{line}(\mathcal{V}_{0},\mathcal{V}_{t}),dist_{traj}(\mathcal{V}_{0},\mathcal{V}_{t}),dist_{step}(\mathcal{V}_{0},\mathcal{V}_{t}))$.
Finally, the observation embeddings
are as follows:
\begin{equation}
\small
\mathcal{O}_{t}=LN(W_{1}^\mathcal{O}[\mathcal{R}_{t}^{'};\mathcal{N}(\mathcal{R}_{t}^{'})])+LN(W_{2}^\mathcal{O}[a_{t};b_{t};c_{t};e_{t}])
\end{equation}
where the $LN$ denotes layer normalization, $W_{1}^\mathcal{O}$ and $W_{2}^\mathcal{O}$ are learnable parameters. A special “stop” token $\mathcal{O}_{t,0}$ is added to $\mathcal{O}_{t}$ for the stop action. 
We use a two-layer transformer to model relations among observation embeddings and output $\mathcal{O}_{t}^{'}$.

\subsubsection{Grid Memory Encoding}\label{sec:instruction relevance_grid_memory_encoding}

As described in Sec.~\ref{sec:grid_memory_mapping}, we need to aggregate multiple grid features in
each cell into one embedding vector. Due to the complexity of the navigation environment, a large number
of grid features within each cell region are not all needed by the agent to complete navigation. The agent needs more critical and highly correlated information with current instruction
to understand the environment. Therefore, we propose an instruction relevance method to aggregate features in each cell. Specifically, for grid features in each
cell $\mathcal{M}_{t,m,n}^{rel}=\{\hat{g}_{t,j}\in\mathcal{\mathbb{R}}^{D}\}_{j=1}^{J}$, where
the corresponding coordinates $\{P^{rel}(\hat{g}_{t,j})\}_{j=1}^{J}$ are all within the cell of row $m$ column
$n$, the number of features in this cell is $J$. We evaluate the relevance of each grid feature to each
token of navigation instruction by computing the relevance matrix
$A$ as:
\begin{equation}
A=(\mathcal{M}_{t,m,n}^{rel}W_{1}^{A})(\mathcal{W}^{'}W_{2}^{A})^{T}
\end{equation}
where $W_{1}^{A}$ and $W_{2}^{A}$ are learnable parameters. After that, we compute the row-wise
max-pooling on $A$ to evaluate the relevance of each grid feature to the instruction as:
\begin{equation}
\alpha_{j}=max(\{A_{j,l}\}_{l=1}^{L})
\end{equation}
At last, we aggregate the grid features within each cell into an embedding vector $E_{t,m,n}$:
\begin{equation}
\eta=softmax(\{\alpha_{j}\}_{j=1}^{J})
\end{equation}
\begin{equation}
E_{t,m,n}=\sum_{j=1}^{J}\eta_{j}(W^{E}\hat{g}_{t,j})
\end{equation}
where $W^{E}$ are learnable parameters.
To represent the spatial relations, we introduce positional information into our grid memory map. Specifically, between each cell center and agent,
we denote the line distance as $q_{t}^{M}$ and represent relative
heading angles as $h_{t}^{M}=(sin\Phi_{t}^{M},cos\Phi_{t}^{M})$.
Then the map features can be obtained:

{\small
\begin{equation}
M_{t}=LN(E_{t})+LN(W^{M}[q_{t}^{M};h_{t}^{M}])
\end{equation}
}
where $W^{M}$ are learnable parameters.

\subsubsection{Navigation Trajectory Encoding}\label{sec:navigation_trajectory_encoding}
In order to implement global action planning, we further introduce
the navigation trajectory into our GridMM. As
shown in Sec.~\ref{rec:observation_encoding}, at time step $t$, the agent receives panoramic features $\mathcal{O}_{t}^{'}$ of waypoint $\mathcal{V}_{t}$.
Then we can obtain visual representation $Avg(\mathcal{O}_{t}^{'})$
of the current waypoint by average pooling of $\mathcal{O}_{t}^{'}$.
As the agent also partially observes candidate waypoints, we use the
view image features $\mathcal{N}(\mathcal{O}_{t}^{'})$ that contains
these navigable waypoints as their visual representation. Between waypoints
and current agent, we denote the line distances as $q^{\mathcal{T}}$,
the relative heading angles as $h_{t}^{\mathcal{T}}=(sin\Phi_{t}^{\mathcal{T}},cos\Phi_{t}^{\mathcal{T}})$,
and the action step embeddings as $u^{\mathcal{T}}$. All historical
waypoint features $\{Avg(\mathcal{O}_{i}^{'})\}_{i=1}^{t-1}$, current
waypoint feature $Avg(\mathcal{O}_{t}^{'})$ and the candidate waypoint
features $\mathcal{N}(\mathcal{O}_{t}^{'})$ form the navigation trajectory:
\begin{align}
\small
\mathcal{T}_{t}&=[\{LN(Avg(\mathcal{O}_{i}^{'}))+LN(W_{1}^{\mathcal{T}}[q_{i}^{\mathcal{T}};h_{i}^{\mathcal{T}}])+u_{i}^{\mathcal{T}}\}_{i=1}^{t};\notag\\&\ \ \ \ \ \ \ LN(\mathcal{N}(\mathcal{O}_{t}^{'}))+LN(W_{2}^{\mathcal{T}}[q_{\mathcal{N}}^{\mathcal{T}};h_{\mathcal{N}}^{\mathcal{T}}])+u_{\mathcal{N}}^{\mathcal{T}}]
\end{align}
where $W_{1}^{\mathcal{T}}$ and $W_{2}^{\mathcal{T}}$ are learnable parameters, a special “stop” token $\mathcal{T}_{t,0}$ is added to $\mathcal{T}_{t}$ for the stop action.

\subsubsection{Cross-modal Reasoning}\label{sec:cross-modal_reasoning}
\begin{figure}
\noindent\begin{minipage}[t]{1\columnwidth}%
\begin{center}
\includegraphics[width=1.0\columnwidth]{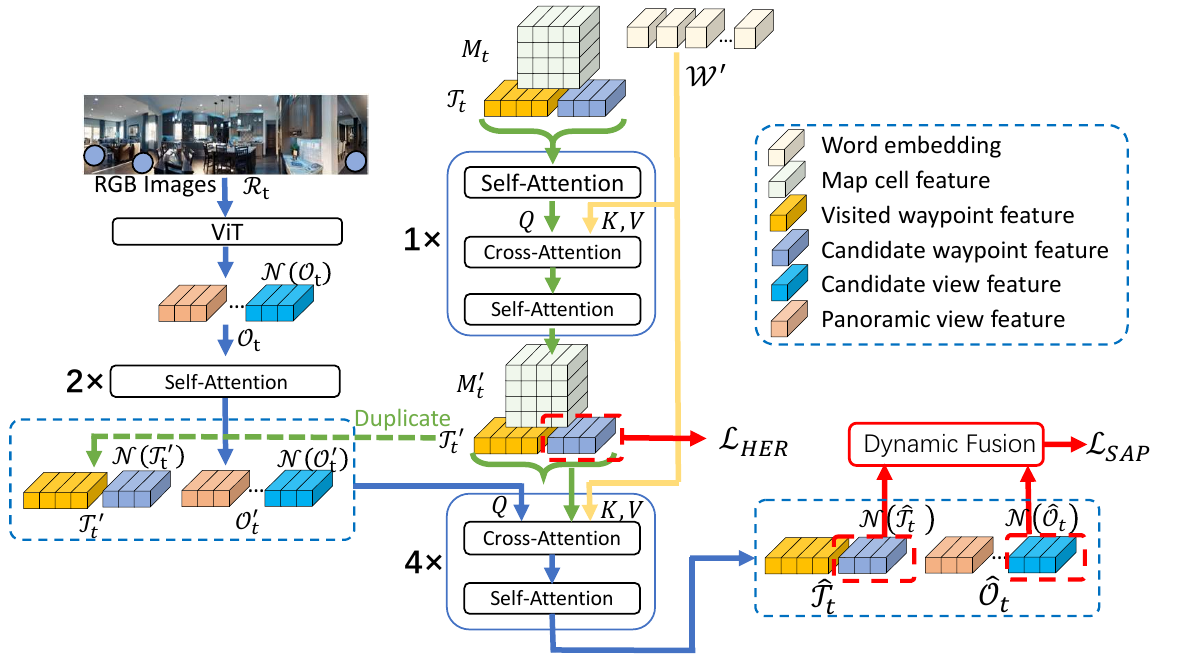}
\par\end{center}%
\end{minipage}
\caption{
The detailed architecture for action prediction.
}
\label{fig:model_details}
\end{figure}

As illustrated in Fig.~\ref{fig:framework}, we concatenate map features and navigation trajectory as $[M_{t};\mathcal{T}_{t}]$, and then use a cross-modal transformer to fuse features from instruction $\mathcal{W}^{'}$ and model space-time
relations, forming the features $[M_{t}^{'};\mathcal{T}_{t}^{'}]$. We specifically design a training loss $\mathcal{L}_{HER}$ (illustrated
in Sec. \ref{sec:training_and_inference}) to supervise this module.

Subsequently, we use another cross-modal transformer with 4 layers to
model vision-language relations and space-time relations. Specifically,
each transformer layer consists of a cross-attention layer and a self-attention
layer. For the cross-attention layer, we input panoramic observation and navigation
trajectory $[\mathcal{O}_{t}^{'};\mathcal{T}_{t}^{'}]$ as queries which
attend over encoded instruction tokens, navigation trajectory and
map features $[\mathcal{W}^{'};\mathcal{T}_{t}^{'}; M_{t}^{'}]$. And then, the self-attention layer takes encoded panoramic observation and navigation
trajectory as input for action reasoning, where the output is denoted as $[\hat{\mathcal{O}_{t}};\mathcal{\hat{T}}_{t}]$.

\subsubsection{Action Prediction}\label{sec:action_prediction}

We predict local navigation scores for the candidate views
$\mathcal{N}(\mathcal{\hat{O}}_{t})$ as below:
\begin{equation}
S_t^{\mathcal{O}}=FFN(\mathcal{N}(\mathcal{\hat{O}}_{t}))
\end{equation}
and predict global navigation scores for the candidate navigable
waypoints $\mathcal{N}(\mathcal{\hat{T}}_{t})$ as below:
\begin{equation}
S_t^{\mathcal{T}}=FFN(\mathcal{N}(\mathcal{\hat{T}}_{t}))
\end{equation}
 where $FFN$ denotes a two-layer feed-forward network.
 To be noted, $S_{t,0}^{\mathcal{O}}$ and $S_{t,0}^{\mathcal{T}}$ are the stop scores. Two separate FFNs are
used to predict local action scores and global action scores, we gated fuse the scores following \cite{chen2022think-GL}:
\begin{equation}
S_t^{fusion}=\lambda_{t}S_t^{\mathcal{O}}+(1-\lambda_{t})S_t^{\mathcal{T}}
\end{equation}
where $\lambda_{t}=sigmoid(FFN([\mathcal{\hat{O}}_{t,0};\mathcal{\hat{T}}_{t,0}]))$.


As illustrated in Fig.~\ref{fig:framework}, $\mathcal{M}_{t}$ is the set of extracted features, and $\mathcal{M}_{t}^{rel}$ is the projected features with relative coordinates. $\mathcal{M}_{t,m,n}^{rel}$  is a subset of $\mathcal{M}_{t}^{rel}$ within a cell, and ${M}_{t}$ is the obtained map features after aggregation. Meanwhile, the detailed architecture for action prediction is illustrated in Fig.~\ref{fig:model_details}. 
For loss functions, $MLM$ and $MVM$ are employed in the same ways as previous works~\cite{chen2022think-GL} (which are omitted in Fig.~\ref{fig:framework} and Fig.~\ref{fig:model_details}). The $SAP$ loss and $HER$ loss are clearly described in Sec.~\ref{sec:training_and_inference}. 
The candidate views are part of the agent's current panoramic observation as candidates for local action prediction. But the candidate waypoints are candidate locations in the global grid map for global action prediction. 
We use ``Dynamic Fusion" to gated fuse these two action scores following DUET~\cite{chen2022think-GL}.

\subsection{Pre-training and   Fine-tuning}\label{sec:training_and_inference}

\paragraph*{Pre-training.}We utilize four tasks to pre-train our model. 

1) Masked language modeling (MLM). We randomly mask out the words
of the instruction with a probability of 15\% and then predict the masked
words $\mathcal{W}_{masked}$.

2) Masked view modeling (MVM). We randomly mask out view
images with a probability of 15\% and predict the semantic labels of masked view images. Similar to \cite{chen2022think-GL}, the target
labels for view images are obtained by an image classification model \cite{2020image-vit} pre-trained on ImageNet.

3) Single-step action prediction (SAP). Given the ground truth action $\mathcal{A}_{t}$, the SAP loss is defined as follows:
\begin{equation}
\mathcal{L}_{SAP}=\sum_{t=1}^{T}CrossEntropy(S_t^{fusion},\mathcal{A}_{t})
\end{equation}

4) Historical environment reasoning (HER). The HER requires the agent to predict the next action only based on the map features and navigation trajectory, without panoramic observations:
\begin{equation}
S_t^{HER}=FFN(\mathcal{N}(\mathcal{T}_{t}^{'}))
\end{equation}
\begin{equation}
\mathcal{L}_{HER}=\sum_{t=1}^{T}CrossEntropy(S_t^{HER},\mathcal{A}_{t})
\end{equation}

\paragraph*{Fine-tuning.}
For fine-tuning, we follow existing works~\cite{chen2022think-GL,Hong2022bridging} to use Dagger~\cite{ross2011reduction} training techniques. Different from the pre-training process which uses the demonstration path, the supervision of fine-tuning
is from a pseudo-interactive demonstrator which selects a navigable
waypoint as the next target with the overall shortest distance from
the current waypoint to the destination. 


\section{Experiment}\label{sec:experiment}

\subsection{Datasets and Evaluation Metrics}
We evaluate our model on the REVERIE~\cite{2020reverie}, R2R~\cite{VLN-2018vision}, SOON~\cite{zhu2021soon} datasets in discrete environments and R2R-CE\ \cite{Krantz2020r2r-ce} in continuous environments. 

\textbf{REVERIE} contains high-level instructions which contain 21 words on average and the path length is between 4 and 7 steps. The predefined object bounding boxes are provided for each panorama, and the agent should select the correct object bounding box from candidates at the end of the navigation path.

\textbf{R2R} provides step-by-step instructions. The average length of instructions is 32 words and the average path length is 6 steps.  

\textbf{SOON} also provides instructions that describe the target locations and target objects. The average length of instructions is 47 words, and the path length is between 2 and 21 steps. However, the object bounding boxes are not provided, the agent needs to predict the center location of the target object. Similar to the settings in~\cite{chen2022think-GL}, we use object detectors~\cite{anderson2018bottom} to obtain candidate object boxes.

\textbf{R2R-CE} are collected based on the
discrete Matterport3D environments\ \cite{matterport3d}, but use the Habitat simulator\ \cite{ramakrishnan2021habitat} to navigate in the continuous environments.

There are several standard metrics\ \cite{VLN-2018vision,2020reverie} in
VLN for evaluating the agent’s performance, including Trajectory Length (TL), Navigation Error (NE), Success Rate
(SR), SR given the Oracle stop policy (OSR), Normalized inverse of the Path Length (SPL), Remote Grounding Success (RGS), and RGS penalized by Path Length (RGSPL). 

\subsection{Implementation Details}
We adopt the pre-trained CLIP-ViT-B/32~\cite{radford2021learning} to extract grid features $G_{t}$ on all datasets. We use the ViT-B/16\ \cite{2020image-vit} pre-trained on ImageNet to extract panoramic view features $\mathcal{R}_{t}^{'}$ on all datasets and extract object features on the REVERIE dataset as it provides bounding boxes. The BUTD object detector~\cite{anderson2018bottom} is utilized on the SOON dataset to extract object bounding boxes. The number of layers for the language encoder, panorama encoder, map and trajectory encoder, and the cross-modal reasoning encoder are  respectively set as 9, 2, 1, and 4 as shown in Fig.~\ref{fig:framework}, all with a hidden size of 768. The parameters of all transformer layers are initialized with the pre-trained LXMERT~\cite{2019-lxmert}.

\subsection{Comparison to State-of-the-Art Methods}
Table \ref{REVERIE_sota},\ref{R2R_sota},\ref{SOON_sota} compare our approach with the previous VLN methods on the REVERIE, R2R and SOON benchmarks. Table~\ref{R2R-CE_sota} compares our approach with the previous VLN-CE methods on the R2R-CE benchmark. Our approach achieves state-of-the-art performance on most metrics, demonstrating the effectiveness of the proposed approach. For the val unseen split of the REVERIE dataset in Table \ref{REVERIE_sota}, our model outperforms the previous DUET \cite{chen2022think-GL} by 4.39\% on SR and 2.74\% on SPL. As shown in Table \ref{R2R_sota} and \ref{SOON_sota}, it also shows performance gains on the R2R and
SOON dataset compared to DUET. In particular, our approach significantly outperforms all previous methods on the R2R-CE dataset in Table \ref{R2R-CE_sota}, demonstrating the 
effectiveness of our GridMM for VLN-CE.

\begin{table*}
\small
\tabcolsep=0.15cm
\centering
{\centering{}}%
\begin{tabular}{c|cccc|cc|cccc|cc}
\hline 
\multirow{3}{*}{Methods} & \multicolumn{6}{c|}{{Val Unseen}} & \multicolumn{6}{c}{{Test Unseen}}\tabularnewline
\cline{2-13} \cline{3-13} \cline{4-13} \cline{5-13} \cline{6-13} \cline{7-13} \cline{8-13} \cline{9-13} \cline{10-13} \cline{11-13} \cline{12-13} \cline{13-13}
 & \multicolumn{4}{c|}{{Navigation}} & \multicolumn{2}{c|}{{Grounding}} & \multicolumn{4}{c|}{{Navigation}} & \multicolumn{2}{c}{{Grounding}}\tabularnewline
\cline{2-13} \cline{3-13} \cline{4-13} \cline{5-13} \cline{6-13} \cline{7-13} \cline{8-13} \cline{9-13} \cline{10-13} \cline{11-13} \cline{12-13} \cline{13-13}
 & {TL\textdownarrow{}} & {OSR\textuparrow{}} & {SR\textuparrow{}} & {SPL\textuparrow{}} & {RGS\textuparrow{}} & {RGSPL\textuparrow{}} & {TL\textdownarrow{}} & {OSR\textuparrow{}} & {SR\textuparrow{}} & {SPL\textuparrow{}} & {RGS\textuparrow{}} & {RGSPL\textuparrow{}}\tabularnewline
\hline 
{VLNBERT\ \cite{hong2021vln-bert}} & {16.78} & {35.02} & {30.67} & {24.90} & {18.77} & {15.27} & {15.68} & {32.91} & {29.61} & {23.99} & {16.50} & {13.51}\tabularnewline
 
{AirBERT\ \cite{2021airbert}} & {18.71} & {34.51} & {27.89} & {21.88} & {18.23} & {14.18} & {17.91} & {34.20} & {30.28} & {23.61} & {16.83} & {13.28}\tabularnewline

{HOP\ \cite{qiao2022HOP}} & {16.46} & {36.24} & {31.78} & {26.11} & {18.85} & {15.73} & {16.38} & {33.06} & {30.17} & {24.34} & {17.69} & {14.34}\tabularnewline

{HAMT\ \cite{chen2021history}} & {14.08} & {36.84} & {32.95} & {30.20} & {18.92} & {17.28} & {13.62} & {33.41} & {30.40} & {26.67} & {14.88} & {13.08}\tabularnewline

{TD-STP\ \cite{zhao2022target}} & {-} & {39.48} & {34.88} & {27.32} & {21.16} & {16.56} & {-} & {40.26} & {35.89} & {27.51} & {19.88} & {15.40}\tabularnewline

{DUET\ \cite{chen2022think-GL}} & {22.11} & {51.07} & {46.98} & {33.73} & {32.15} & {23.03} & {21.30} & {56.91} & {52.51} & {36.06} & {31.88} & {22.06}\tabularnewline

{BEVBert\ \cite{Dong2022bevbert}} & {-} & {56.40} & \textbf{51.78} & {36.37} & \textbf{34.71} & {24.44} & {-} & {57.26} & {52.81} & {36.41} & {32.06} & {22.09}\tabularnewline

\hline 
GridMM (Ours) & {23.20} & \textbf{57.48} & {51.37} & \textbf{36.47} & {34.57} & \textbf{24.56} & {19.97} & \textbf{59.55} & \textbf{53.13} & \textbf{36.60} & \textbf{34.87} & \textbf{23.45}\tabularnewline
\hline 
\end{tabular}
\vspace{2pt}
\caption{Evaluation on the REVERIE dataset.}\label{REVERIE_sota}
\end{table*}

\begin{table}
\small
\tabcolsep=0.05cm
\centering{}%
\begin{tabular}{c|cccc|cccc}
\hline 
\multirow{2}{*}{Methods} & \multicolumn{4}{c|}{Val Unseen} & \multicolumn{4}{c}{Test Unseen}\tabularnewline
\cline{2-9} \cline{3-9} \cline{4-9} \cline{5-9} \cline{6-9} \cline{7-9} \cline{8-9} \cline{9-9}
& TL\textdownarrow{} & NE\textdownarrow{} & SR\textuparrow{} & SPL\textuparrow{} & TL\textdownarrow{} & NE\textdownarrow{} & SR\textuparrow{} & SPL\textuparrow{}\tabularnewline
\hline 
VLNBERT\ \cite{hong2021vln-bert} & 12.01 & 3.93 & 63 & 57 & 12.35 & 4.09 & 63 & 57\tabularnewline

AirBERT\ \cite{2021airbert} & 11.78 & 4.01 & 62 & 56 & 12.41 & 4.13 & 62 & 57\tabularnewline

SEvol\ \cite{Chen_2022_Reinforced} & 12.26 & 3.99 & 62 & 57 & 13.40 & 4.13 & 62 & 57\tabularnewline
 
HOP\ \cite{qiao2022HOP} & 12.27 & 3.80 & 64 & 57 & 12.68 & 3.83 & 64 & 59\tabularnewline
 
HAMT\ \cite{chen2021history} & 11.46 & 2.29 & 66 & 61 & 12.27 & 3.93  & 65 & 60\tabularnewline
 
TD-STP\ \cite{zhao2022target} & - & 3.22 & 70 & 63 & - & 3.73 & 67 & 61\tabularnewline

DUET\ \cite{chen2022think-GL} & 13.94 & 3.31 & 72 & 60 & 14.73 & 3.65 & 69 & 59\tabularnewline
 
BEVBert\ \cite{Dong2022bevbert} & 14.55 & \textbf{2.81}  & \textbf{75} & \textbf{64} & 15.87 & \textbf{3.13} & \textbf{73} & \textbf{62}\tabularnewline
\hline

GridMM (Ours) & 13.27 & 2.83 & \textbf{75} & \textbf{64} & 14.43 & 3.35 & \textbf{73} & \textbf{62} \tabularnewline

\hline 
\end{tabular}
\vspace{2pt}
\caption{Evaluation on the R2R dataset.}\label{R2R_sota}
\end{table}

\begin{table}
\small
\tabcolsep=0.04cm
\centering{}%
\begin{tabular}{c|c|ccccc}
\hline 
Split & Method & TL\textdownarrow{} & OSR\textuparrow{} & SR\textuparrow{} & SPL\textuparrow{} & RGSPL\textuparrow{}\tabularnewline\cline{2-7} \cline{3-7} \cline{4-7} \cline{5-7} \cline{6-7} \cline{7-7}
\hline 
\multirow{3}{*}{Val Unseen} & GBE\ \cite{zhu2021soon} & 28.96 & 28.54 & 19.52 & 13.34 & 1.16\tabularnewline
& DUET\ \cite{chen2022think-GL} & 36.20 & 50.91 & 36.28 & 22.58 & 3.75\tabularnewline
& GridMM (Ours) & 38.92 & \textbf{53.39} & \textbf{37.46} & \textbf{24.81} & \textbf{3.91}\tabularnewline
\hline 
\multirow{3}{*}{Test Unseen} & GBE\ \cite{zhu2021soon} & 27.88 & 21.45 &  12.90 & 9.23 & 0.45
\tabularnewline
& DUET\ \cite{chen2022think-GL} & 41.83 & 43.00 & 33.44 &  \textbf{21.42} &  \textbf{4.17}\tabularnewline
& GridMM (Ours) & 46.20 & \textbf{48.02} & \textbf{36.27} & 21.25 & 4.15\tabularnewline
\hline 
\end{tabular}
\vspace{1pt}
\caption{Evaluation on the SOON dataset.}\label{SOON_sota}
\end{table}

\begin{table*}
\small
\tabcolsep=0.1cm
\centering{}%
\begin{tabular}{c|ccccc|ccccc|ccccc}
\hline 
\multirow{2}{*}{Methods} & \multicolumn{5}{c|}{Val Seen} & \multicolumn{5}{c|}{Val Unseen} & \multicolumn{5}{c}{Test Unseen}\tabularnewline
\cline{2-16} \cline{3-16} \cline{4-16} \cline{5-16} \cline{6-16} \cline{7-16} \cline{8-16} \cline{9-16} \cline{10-16} \cline{11-16} \cline{12-16} \cline{13-16} \cline{14-16} \cline{15-16} \cline{16-16}
 & TL\textdownarrow{} & NE\textdownarrow{} & OSR\textuparrow{} & SR\textuparrow{} & SPL\textuparrow{} & TL\textdownarrow{} & NE\textdownarrow{} & OSR\textuparrow{} & SR\textuparrow{} & SPL\textuparrow{} & TL\textdownarrow{} & NE\textdownarrow{} & OSR\textuparrow{} & SR\textuparrow{} & SPL\textuparrow{}\tabularnewline
\hline 
VLN-CE\textcolor{black}{${\color{red}^{{\color{black}\ast}}}$}\ \cite{Krantz2020r2r-ce} & 9.26 & 7.12 & 46 & 37 & 35 & 8.64 & 7.37 & 40 & 32 & 30 & 8.85 & 7.91 & 36 & 28 & 25\tabularnewline
 
AG-CMTP\ \cite{chen2021topo} & - & 6.60 & 56.2 & 35.9 & 30.5 & - & 7.9 & 39.2 &23.1 & 19.1 & - & - & - & - & -\tabularnewline

R2R-CMTP\ \cite{chen2021topo} & - & 7.10 & 45.4 & 36.1 & 31.2 & - & 7.9 & 38.0 & 26.4 & 22.7 & - & - & - & - & -\tabularnewline

WPN\ \cite{Krantz2021waypointmodel} & 8.54 & 5.48 & 53 & 46 & 43 & 7.62 & 6.31 & 40 & 36 & 34 & 8.02 & 6.65 & 37 & 32 & 30\tabularnewline

LAW\textcolor{black}{${\color{red}^{{\color{black}\ast}}}$}\ \cite{Raychaudhuri2021law} & 9.34 & 6.35 & 49 & 40 & 37 & 8.89 & 6.83 & 44 & 35 & 31 & - & - & - & - & -\tabularnewline

CM$^{2}$\textcolor{black}{${\color{red}^{{\color{black}\ast}}}$}\ \cite{georgakis2022cm2} & 12.05 & 6.10 & 50.7 & 42.9 & 34.8 & 11.54 & 7.02 & 41.5 & 34.3 & 27.6  
 & 13.9 & 7.7 & 39 & 31 & 24\tabularnewline

CM$^{2}$-GT\textcolor{black}{${\color{red}^{{\color{black}\ast}}}$}\ \cite{georgakis2022cm2} & 12.60 & 4.81 & 58.3 & 52.8 & 41.8 & 10.68 & 6.23 & 41.3 & 37.0 & 30.6 & - & - & - & - & -\tabularnewline

WS-MGMap\textcolor{black}{${\color{red}^{{\color{black}\ast}}}$}\ \cite{chen2022weakly} & 10.12 & 5.65 & 51.7 & 46.9 & 43.4 & 10.00 & 6.28 & 47.6 & 38.9 & 34.3 & 12.30 & 7.11 & 45 & 35 & 28\tabularnewline

Sim-2-Sim\ \cite{krantz2022sim2sim} & 11.18 & 4.67 & 61 & 52 & 44 & 10.69 & 6.07 & 52 & 43 & 36 & 11.43 & 6.17 & 52 & 44 & 37\tabularnewline

ERG\textcolor{black}{${\color{red}^{{\color{black}\dagger}}}$}\ \cite{Ting2023graph-vlnce} & 11.8 & 5.04 & 61 & 46 & 42 & 9.96 & 6.20 & 48 & 39 & 35 & - & - & - & - & -\tabularnewline

CMA\textcolor{black}{${\color{red}^{{\color{black}\dagger}}}$} \ \cite{Hong2022bridging} & 11.47 & 5.20 & 61 & 51 & 45 & 10.90 & 6.20 & 52 & 41 & 36 & 11.85 & 6.30 & 49 & 38 & 33\tabularnewline

VLNBERT\textcolor{black}{${\color{red}^{{\color{black}\dagger}}}$}\ \cite{Hong2022bridging} & 12.50 & 5.02 & 59 & 50 & 44 & 12.23 & 5.74 & 53 & 44 & 39 & 13.31 & 5.89 & 51 & 42 & 36\tabularnewline

\hline 

DUET\textcolor{black}{${\color{red}^{{\color{black}\dagger}}}$} (Ours)\ \cite{chen2022think-GL} & 12.62 & \textbf{4.13} & 67 & 57 & 49 & 13.04 & 5.26 & 58 & 47 & 39 & 13.13 & 5.82 & 50 & 42 & 36\tabularnewline

GridMM\textcolor{black}{${\color{red}^{{\color{black}\dagger}}}$} (Ours) & 12.69 & 4.21 & \textbf{69} & \textbf{59} & \textbf{51} & 13.36 & \textbf{5.11} & \textbf{61} & \textbf{49} & \textbf{41} & 13.31 & \textbf{5.64} & \textbf{56} & \textbf{46} & \textbf{39}
 \tabularnewline
\hline 
\end{tabular}
\vspace{2pt}
\caption{Evaluation on the R2R-CE dataset. The methods marked with \textcolor{black}{${\color{red}^{{\color{black}\ast}}}$} use a forward-facing camera with a 90-degree HFOV instead of panoramic images. The methods marked with \textcolor{black}{${\color{red}^{{\color{black}\dagger}}}$} use the same waypoint predictor\ \cite{Hong2022bridging} for a fair comparison. Especially, we transfer the pre-trained DUET model to R2R-CE.}\label{R2R-CE_sota}
\end{table*}

\begin{table}
\small
\tabcolsep=0.08cm
\centering{}%
\begin{tabular}{c|ccccc}
\hline 
Mapping methods & TL\textdownarrow{} & NE\textdownarrow{} & OSR\textuparrow{} & SR\textuparrow{} & SPL\textuparrow{}\tabularnewline
\hline 
No Map & 14.61 & 5.64 & 57.24 & 45.19 & 37.82\tabularnewline

DUET (topological map) & 13.04 & 5.26 & 57.91 & 47.02 & 38.86\tabularnewline

Top-down semantic map & 13.78 & 5.33 & 57.46 & 46.36 & 38.41\tabularnewline
 
Map with object features & 13.15 & 5.39 & 59.12 & 47.61 & 40.13\tabularnewline

Our GridMM & 13.36 & \textbf{5.11} & \textbf{60.90} & \textbf{49.05} & \textbf{40.99}\tabularnewline
\hline 
\end{tabular}
\vspace{3pt}
\caption{Comparison among different maps on the val unseen split of the R2R-CE dataset. Row 1 is our baseline method that uses our proposed model but without grid features. Row 3 takes top-down semantic maps as substitutes for grid features. Row 4 takes object features extracted from a detection model~\cite{zhang2021vinvl} as substitutes for grid features. Row 5 is our proposed GridMM. 
} \label{map_comparison}
\end{table}

\begin{center}
\begin{table}
\small
\noindent\begin{minipage}[t]{1\columnwidth}%
\tabcolsep=0.08cm
\begin{center}
\begin{tabular}{cccc|ccccc}
\hline 
GridMM & Ego. & Traj. & Instr. & TL\textdownarrow{} & NE\textdownarrow{} & OSR\textuparrow{} & SR\textuparrow{} & SPL\textuparrow{}\tabularnewline
\hline 
  & $\checkmark$ & $\checkmark$ & & 14.61 & 5.64 & 57.24 & 45.19 & 37.82
\tabularnewline

$\checkmark$ &  &$\checkmark$ &$\checkmark$ & 13.24 & 5.23 & 59.11 & 48.72
 & 40.14\tabularnewline
 
$\checkmark$ & $\checkmark$ &  &$\checkmark$& 13.14 & 5.24 & 58.35 & 47.42
 & 39.41\tabularnewline

$\checkmark$ & $\checkmark$ & $\checkmark$ &  & 13.22 & 5.39 & 59.75 & 48.63 & 39.83\tabularnewline
\hline 
$\checkmark$ & $\checkmark$ & $\checkmark$ &$\checkmark$& 13.36 & \textbf{5.11} & \textbf{60.90} & \textbf{49.05} & \textbf{40.99}\tabularnewline
\hline 
\end{tabular}
\par\end{center}%
\end{minipage}
\vspace{1pt}
\caption{Ablation study results on the val unseen split of R2R-CE dataset. “GridMM” denotes using grid memory map. “Ego.” denotes using the egocentric view shown in Fig.\ \ref{fig:map_grow}(b) instead of the map in Fig.\ \ref{fig:map_grow}(a). “Traj.” denotes using trajectory memory and features. “Instr.” denotes using instruction relevance aggregation method instead of average pooling grid features in each cell.}\label{ablation_study}
\end{table}
\par\end{center}

\begin{table}
\small
\tabcolsep=0.1cm
\centering{}%
\begin{tabular}{c|ccccc}
\hline 
Map scale & TL\textdownarrow{} & NE\textdownarrow{} & OSR\textuparrow{} & SR\textuparrow{} & SPL\textuparrow{}\tabularnewline
\hline 
8$\times$8 & 13.42 & 5.23 & 58.58 & 47.07 & 39.49\tabularnewline

14$\times$14 & 13.36 & 5.11 & \textbf{60.90} & 49.05 & 40.99\tabularnewline

20$\times$20 & 12.59 & \textbf{4.95} & 57.86 & \textbf{49.86} & \textbf{42.52}\tabularnewline

\hline
\end{tabular}
\vspace{4pt}
\caption{The effect of different map scales on the val unseen split of the R2R-CE dataset.}\label{map_scale}
\end{table}

\subsection{Ablation Study}
We compare the performance of different maps representing the visited environments on the val unseen split of the R2R-CE dataset.
\paragraph*{1) Grid memory map vs. other maps.} 
As shown in Table \ref{map_comparison}, we compare the effects of three different maps on the R2R-CE dataset. For row 2, we followed the same model structure as\ \cite{chen2022think-GL}. For row 3, we take the top-down semantic map as a substitute for grid features. Specifically, we followed CM$^{2}$\ \cite{georgakis2022cm2} to obtain an egocentric top-down semantic map, and use a convolution layer to extract semantic features in each cell instead of grid features. Row 4 uses a pre-trained object detection model VinVL\ \cite{zhang2021vinvl} to detect multiple objects and extract their features as substitutes for grid features. More detailed experimental setups can be found in the supplementary materials.

In Table~\ref{map_comparison}, all results with maps (rows 2-5) are better than the baseline method (row 1), which fully demonstrates the necessity of constructing maps representing the environments for VLN. Furthermore, our GridMM is better than DUET (topological map), as GridMM contains more fine-grained information. The method with a top-down semantic map (row 3) is beneficial to navigation, but it is still inferior to row 4, row 5, and even row 2 with the topological map. The reason is that 
map features extracted from the semantic map have a large gap with panoramic visual features. 
Results in Table~\ref{map_comparison} indicate that GridMM is superior to the topological and semantic maps.

\paragraph*{2) Grid features vs. object features.}
By comparing the results of row 4 and row 5 in Table~\ref{map_comparison}, we can find out that the grid map using grid features works better than using object features. This  is mainly because of the following reasons: (i) Object features from object detection model~\cite{zhang2021vinvl} are not enough to represent all visual information, such as house structure and background. (ii) Grid features from CLIP~\cite{radford2021learning} have larger semantic space and better generalization ability. Different from previous methods~\cite{Chen_2022_Reinforced}~\cite{Ting2023graph-vlnce} of obtaining environment representation based on objects, grid features are of great importance for representing environments.
\vspace{-3ex}
\paragraph*{3) Is it necessary that map in an egocentric view?}
As shown in Sec.~\ref{sec:grid_memory_mapping} and Fig. \ref{fig:map_grow}, we discussed two coordinate systems for our grid memory map, \textit{i.e.}, absolute coordinates and dynamically relative coordinates. Row 2 in Table \ref{ablation_study} shows the results of the absolute coordinate system, where the results are obtained by removing the process of coordinate transformation (\textit{i.e.}, depicted in Equation~\ref{formula:relative_coordinates}) but the side  length $L_t$ of the map increases with the expansion of the visited environment (\textit{i.e.}, depicted in Equation~\ref{formula:side_length}). For the settings in row 2, $q_{t}^{M}$ and $h_{t}^{M}$ (\textit{i.e.}, depicted in Sec. \ref{sec:instruction relevance_grid_memory_encoding}) are replaced with the line distance and heading angle between each cell center and the start waypoint. The experimental results show that the egocentric relative coordinate system works better than the absolute coordinate system. It is mainly because maps with the absolute coordinate are not efficient enough to align the candidate observations and the instruction.

\paragraph*{4) The effect of navigation trajectory information.}
As illustrated in Table \ref{ablation_study}, row 3 is inferior to row 5. The results verify the necessity of navigation trajectory, which helps with instruction grounding. The hypothesis is that the navigation trajectory can provide information for grounding the next step to \textit{“cross in front of the refrigerator”} or to \textit{“walk past the wood table and chairs on your right”}, as illustrated in Fig.~\ref{fig:introduction} (c).
\paragraph*{5) The effect of instruction relevance aggregation method.} 
As shown in Table \ref{ablation_study}, row 5 with instruction relevance aggregation has better performance than row 4. Row 4 simply aggregates features in each map cell via average pooling, which makes it difficult to dig out critical visual cues. Our aggregation method evaluates the relevance of each grid feature to navigation instruction and uses the attention mechanism to filter out irrelevant features and capture critical clues.
\paragraph*{6) The effect of map scale.}
As shown in Table \ref{map_scale}, we evaluate
the scale of our GridMM. We observe an upward trend in navigation performance as the map scale increases. This is mainly because a map with a larger scale can accommodate more environmental details and represent spatial relations more precisely. However, increasing the map scale leads to heavy computational cost but the gains are slight. So we choose a relatively balanced scale (\textit{i.e.}, 14$\times$14).

\subsection{Statistical Analyses}\label{sec:sup_statistical_analyses}
\paragraph{The side length of the GridMM.}
\begin{figure}[ht]
\noindent\begin{minipage}{1\columnwidth}%
\begin{center}
\includegraphics[width=0.86\columnwidth]{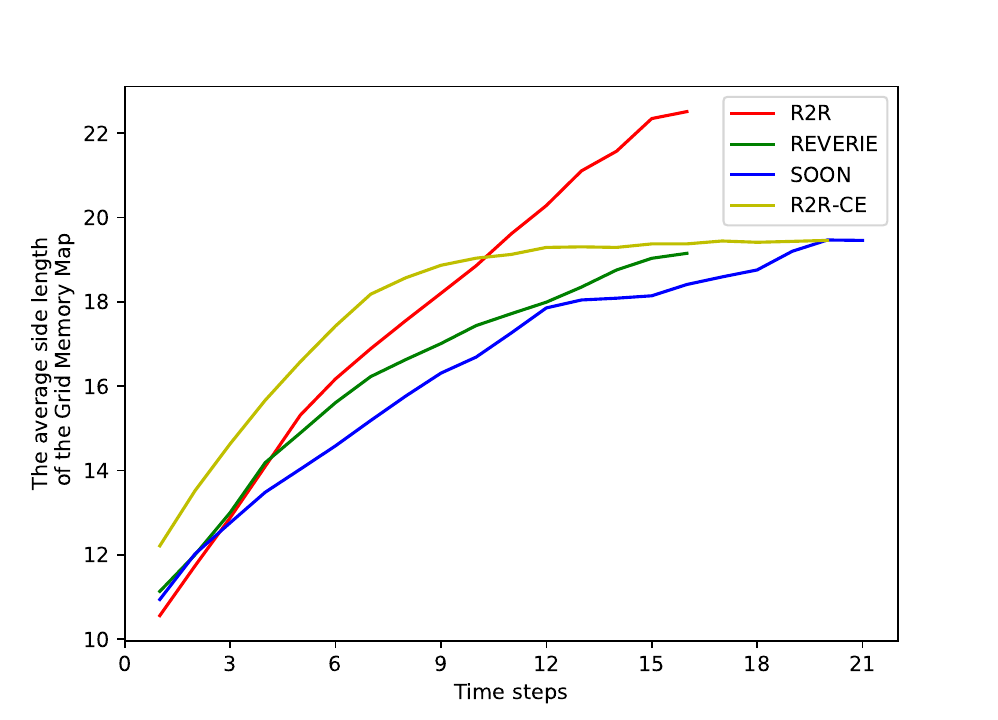}
\end{center}%
\end{minipage}
\caption{The side length of the Grid Memory Map increases with the time steps. The x-axis indicates the 
time steps, and the y-axis is the average side length (meters) of the Grid Memory Map.
}\label{fig:side_length}
\end{figure}

\begin{figure}[ht]
\noindent\begin{minipage}{1\columnwidth}%
\begin{center}
\includegraphics[width=0.86\columnwidth]{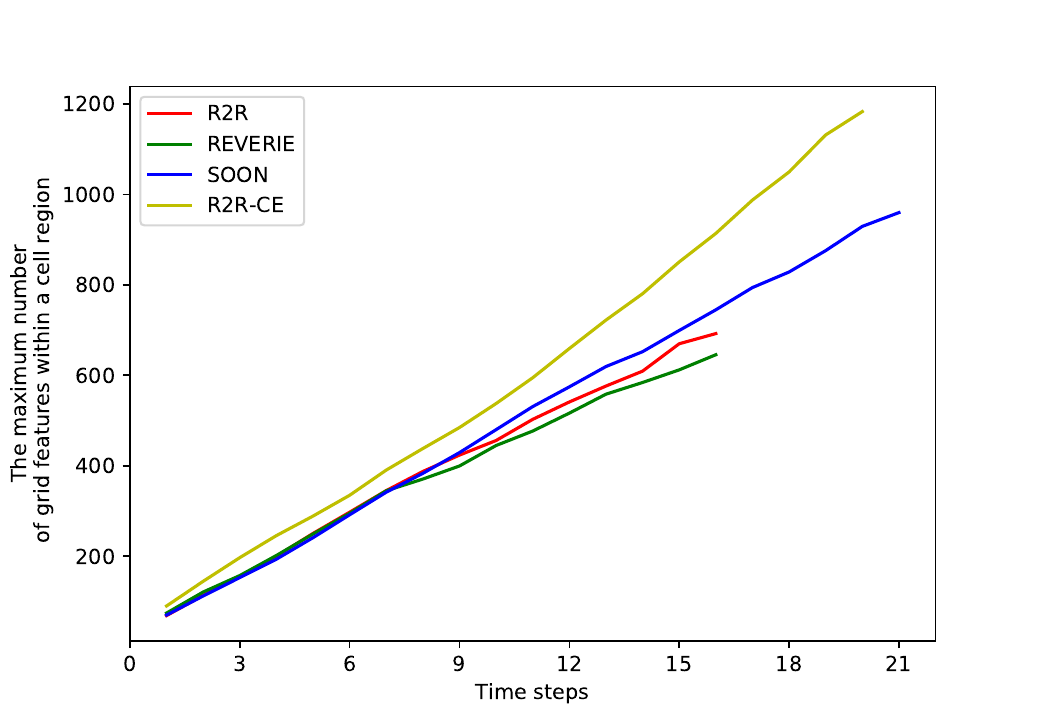}
\end{center}%
\end{minipage}
\caption{The maximum number of grid features within a cell region increases with the time steps. The x-axis indicates the navigation time steps, and the y-axis is the maximum number of grid features within a cell region.
}\label{fig:number_of_grid_fts}
\end{figure}

As illustrated in Fig.~\ref{fig:side_length}, the side length of GridMM increases with the expansion of the visited environment. For all datasets, the side length gradually increases from about 10 meters to about 20 meters during navigation. Obviously, the fixed-size map is difficult to adapt to the visited environment that constantly expands, thus our GridMM with a dynamically relative coordinate system works better. Compared with other datasets, R2R has a larger map size at the end of navigation. It shows that the agent can explore new unvisited environments more on the R2R dataset.

\paragraph{The number of grid features within each cell region.}

As illustrated in Fig.~\ref{fig:number_of_grid_fts}, the maximum number of grid features within a cell region exceeds 600 at the end of navigation on all datasets. A large number
of grid features within a cell region contain noise and are redundant. The average pooling of so many features is not efficient enough, resulting in critical cues being overwhelmed by noise. In contrast, the instruction relevance aggregation method works better than the average pooling, which filters out irrelevant features and captures critical clues.

\section{Conclusion}

In this paper, we propose a top-down egocentric and dynamically growing Grid Memory Map (\textit{i.e.}, GridMM) to structure the visited environment for VLN. Moreover, an instruction relevance aggregation module is proposed to capture fine-grained visual clues relevant to instructions. We comprehensively analyze the effectiveness of our model and compare it with other methods. Our GridMM provides both global space-time perception and local detailed clues, thus enabling more accurate navigation results. However, there are still some limitations to our approach, regarding how to handle multi-floor environments remains open. In the future, we will continuously explore how to better represent the indoor environment for VLN and Embodied AI.

\paragraph*{Acknowledgment.} This work was supported in part by
the National Natural Science Foundation of China under
Grants 62125207, 62102400, 62272436, and U1936203, in part by the National Postdoctoral Program
for Innovative Talents under Grant BX20200338.

\clearpage
{\small
\bibliographystyle{ieee_fullname}
\bibliography{egbib}
}

\appendix
\section*{Appendix}

\section{Datasets}\label{sec:sup_datasets}
We 
evaluate our approach in discrete environments (\textit{e.g.}, R2R~\cite{VLN-2018vision}, REVERIE~\cite{2020reverie}, and SOON~\cite{zhu2021soon}), and further analyze many characteristics of our approach in continuous environments (\textit{e.g.}, R2R-CE~\cite{Krantz2020r2r-ce} and RxR-CE~\cite{2020-RXR}).

All the benchmarks in discrete environments build upon the Matterport3D environment~\cite{matterport3d} and contain 90 photo-realistic houses. Each house contains a set of navigable locations, and each location is represented by the corresponding panorama image and GPS coordinates. We adopt the standard split of houses into training, val seen, val unseen, and test splits. Houses in the val seen split are the same as in training, while houses in val unseen and test splits are different from training. All splits in discrete environments are consistent with Chen \textit{et al.}~\cite{chen2022think-GL}.

R2R-CE~\cite{Krantz2020r2r-ce} transfers the discrete paths in R2R dataset to continuous trajectories on the Habitat simulator~\cite{2019habitat}. 
RxR-CE~\cite{2020-RXR} transfers the discrete paths in RxR dataset to continuous trajectories
on the Habitat simulator~\cite{2019habitat}.

\section{Performance in RxR-CE}\label{sec:rxr_ce}

\begin{table}[h]
\small
\tabcolsep=0.04cm
\centering{}%
\begin{tabular}{c|c|ccccc}
\hline 
Method & TL & NE\textdownarrow{} & SR\textuparrow{} & SPL\textuparrow{} & nDTW\textuparrow{} & SDTW\textuparrow{}\tabularnewline\cline{2-6} \cline{3-6} \cline{4-6} \cline{5-6} \cline{6-6}
\hline 
VLN-CE~\cite{Krantz2020r2r-ce} & 7.33 & 12.1 & 13.93 & 11.96 & 30.86 & 11.01\tabularnewline
CMA~\cite{Hong2022bridging} & 20.04 & 10.4 & 24.08 & 19.07 & 37.39 & 18.65\tabularnewline
VLNBERT~\cite{Hong2022bridging} & 20.09 & 10.4 & 24.85 & 19.61 & 37.30 & 19.05\tabularnewline
DUET~\cite{chen2022think-GL}(Ours) & 21.48 & 9.78 & 29.93 & 23.12 & 42.46 & 25.39\tabularnewline
GridMM (Ours) & 21.13 & \textbf{8.42} & \textbf{36.26} & \textbf{30.14} & \textbf{48.17} & \textbf{33.65}\tabularnewline
\hline 
\end{tabular}
\vspace{3pt}
\caption{Evaluation on the test unseen split of RxR-CE dataset.}\label{rxr_ce_sota}
\end{table}

As shown in Table~\ref{rxr_ce_sota}, our GridMM achieves competitive results on longer trajectory navigation such as RxR-CE.

\section{Experimental Details}\label{sec:sup_experimental_setups}
\subsection{Training Details}

For the REVERIE dataset, we combine the original dataset with augmented data synthesized by DUET~\cite{chen2022think-GL} to pre-train our model with a batch size of 32 and a learning rate of 5e-5 for 100k iterations, using 3 NVIDIA RTX3090 GPUs. Then we fine-tune it with the batch size of 4 and a learning rate of 1e-5 for 50k iterations on 3 GPUs. 

For the SOON dataset, we only use the original data with automatically cleaned object bounding boxes, sharing the same settings with DUET~\cite{chen2022think-GL}. We pre-train the model with a batch size of 16 and a learning rate of 5e-5 for 40k iterations using 3 GPUs, and then fine-tune it with a batch size of 2 and a learning rate of 5e-5 for 20k iterations on 3 GPUs. 

For the R2R dataset, additional augmented data in\ \cite{hao2020towards} is used for pre-training following DUET~\cite{chen2022think-GL}. Using 3 GPUs, we pre-train our model with a batch size of 32 and a learning rate of 5e-5 for 100k iterations. Then we fine-tune it with the batch size of 4 and a learning rate of 1e-5 for 50k iterations on 3 GPUs.

For the R2R-CE dataset, we transfer the model pre-trained on the R2R dataset to continuous environments, and fine-tune it with a batch size of 8 and a learning rate of 1e-5 for 30 epochs using 3 RTX3090 GPUs.

For all the datasets, the best model is selected by SPL on the val unseen split.

\subsection{Ablation Details}
\paragraph{Top-down semantic map.}

For row 3 in Table~\ref{map_comparison}, we follow CM$^{2}$\ \cite{georgakis2022cm2} to obtain a $448$$\times$$448$ top-down semantic map. Specifically, we use a pre-trained UNet~\cite{unet} from CM$^{2}$~\cite{georgakis2022cm2} to produce semantic segmentation of observation images, and then project pixels into a unified top-down semantic map. After dividing the top-down semantic map into multiple patches with a scale of 32$\times$32, a convolution layer is used to encode these patches into embeddings with a hidden size of 768. We take these semantic embeddings as the map features.

\paragraph{Map with object features.}
For row 4 in Table~\ref{map_comparison}, a pre-trained detection model VinVL~\cite{zhang2021vinvl} is utilized to detect multiple objects in each view image, and then we take 10 object features with the highest confidence score as substitutes for grid features. For the coordinate of each object, it is obtained via the center point of the bounding box.

\vspace{0.4in}
\section{Analysis of Computational Cost}\label{sec:computational_cost_analyses}

Referring to~\cite{efficient_llm}, we describe how we calculate the number of Floating-point Operations (FLOPs) in VLN models as follows:

1) Matrix multiplication ($A_{m\times k}\times B_{k\times n}$): 
\begin{center}
\vspace{-0.05in}
$2mkn$ FLOPs
\end{center}

2) 2-layer MLP (sequence length $s$, increase the hidden size to $4h$ and then reduces it back to $h$):
\begin{center}
\vspace{-0.05in}
$16sh^2$ FLOPs
\end{center}

3) Self-attantion block (sequence length $s$, hidden size $h$):
\begin{center}
\vspace{-0.05in}
$4s^2h+8sh^2$ FLOPs
\end{center}

4) Cross-attantion block (query sequence length $s$, key and value sequence length $t$, hidden size $h$):
\begin{center}
\vspace{-0.05in}
$4sh^2+4th^2+4sth$ FLOPs
\end{center}

\begin{figure}[h]
    \centering
    {
        \includegraphics[width=2.8in]{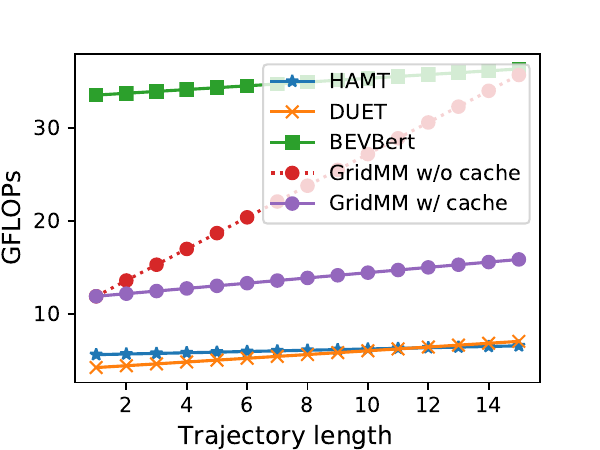}}
    {
    \caption{\small{GFLOPs at different trajectory lengths keeping instruction length as 32. The computational cost of visual encoders and text encoders is omitted for a more intuitive comparison.}}
    \label{reb_fig.1}}
\end{figure}

\begin{figure}[h]
    \centering
    {
        \includegraphics[width=2.8in]{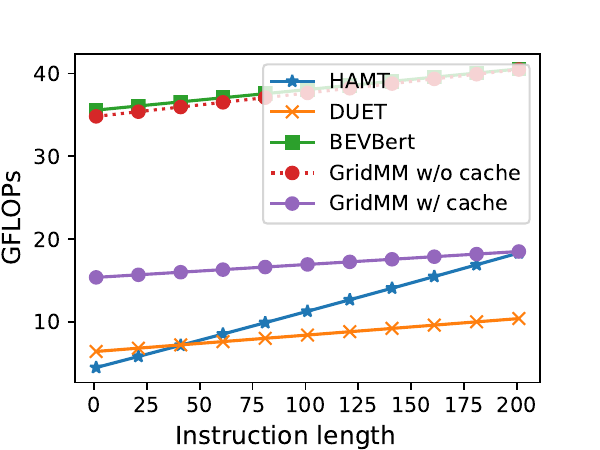}}
    {
    \caption{\small{GFLOPs with different instruction lengths keeping trajectory length as 15. The computational cost of visual encoders and text encoders is omitted for a more intuitive comparison.}}
    \label{reb_fig.2}}
\end{figure}

We calculate GFLOPs (Giga Floating-point Operations) on the R2R dataset, as illustrated in Fig.~\ref{reb_fig.1} and Fig.~\ref{reb_fig.2}. ``GridMM w/o cache" denotes that our GridMM updates each cell of the grid map in all navigation steps without any cache. By using the cache (which stores previous results for later use), the computational cost is significantly reduced. For the same grid features in all navigation steps, during updating the cells of the grid map, we only need to recompute the
positions of grid features, without recomputing their relevance value in the relevance matrix with the instruction. The reason is that, for Equations (6) and (9), the outputs of $\hat{g}_{t,j}W_{1}^{A}$ (where $\hat{g}_{t,j}$ is a part of $\mathcal{M}_{t,m,n}^{rel}$), $\mathcal{W}^{'}W_{2}^{A}$ and $W^{E}\hat{g}_{t,j}$ in all navigation steps is the same and can be cached for reuse. GFLOPs of ``GridMM w/ cache" are significantly lower than that of BEVBert~\cite{Dong2022bevbert}.
During attention computation, the number of metric map features in BEVBert exceeds 400, introducing a huge computational cost. However, the number of map features in GridMM is less than 200 and they are only used as key and value tokens in cross-attention computation, which greatly reduces the computational cost.

\end{document}